\pdfoutput=1

\documentclass[11pt]{article}

\usepackage{EMNLP2022}

\usepackage{times}
\usepackage{latexsym}

\usepackage[T1]{fontenc}

\usepackage[utf8]{inputenc}

\usepackage{microtype}

\usepackage{inconsolata}

\usepackage{multirow}
\usepackage{booktabs}
\usepackage{pifont}
\usepackage[frozencache,cachedir=minted-cache]{minted}

\title{Considerations for meaningful sign language machine translation based on glosses}

\author{Mathias M\"{u}ller$^1$, Zifan Jiang$^1$, Amit Moryossef$^{1,2,3}$, Annette Rios$^1$ and Sarah Ebling$^1$ \\
  $^1$ Department of Computational Linguistics, University of Zurich, Switzerland \\
  $^2$ ETH Zurich, Switzerland, $^3$ Bar-Ilan University, Tel Aviv, Israel \\
  \texttt{\{mmueller,jiang,rios,ebling\}@cl.uzh.ch}, \texttt{amitmoryossef@gmail.com}}

\begin{document}
\maketitle
\begin{abstract}

Automatic sign language processing is gaining popularity in Natural Language Processing (NLP) research \citep{yin-etal-2021-including}. In machine translation (MT) in particular, sign language translation based on \textit{glosses} is a prominent approach.
In this paper, we review recent works on neural gloss translation. We find that limitations of glosses in general and limitations of specific datasets are not discussed in a transparent manner and that there is no common standard for evaluation.

To address these issues, we put forward concrete recommendations for future research on gloss translation. Our suggestions advocate awareness of the inherent limitations of gloss-based approaches, realistic datasets, stronger baselines and convincing evaluation.

\end{abstract}

\section{Introduction}
\label{sec:intro}

Automatic sign language processing is becoming more popular in Natural Language Processing (NLP) research \citep{yin-etal-2021-including}. In machine translation (MT) in particular, many recent publications have proposed sign language translation based on \textit{glosses}. Glosses provide semantic labels for individual signs. They typically consist of the base form of a word in the surrounding spoken language written in capital letters (see Table \ref{tbl:gloss_example}).
Even though glosses are not a complete representation of signs (see e.g. \citealt{pizzuto:06001:sign-lang:lrec}), they are often adopted in MT because, by virtue of being textual, they fit seamlessly into existing MT pipelines and existing methods seemingly require the least modification.

In this paper, we review recent works on neural gloss translation. We find that limitations of gloss-based approaches in general and limitations of specific datasets are not transparently discussed as inherent shortcomings. Furthermore, among gloss translation papers there is no common standard for evaluation, especially regarding the exact method to compute BLEU scores.


Experiments in sign language translation should be informed by sign language expertise and should be performed according to the best practices already established in the MT community.

To alleviate these problems going forward, we make practical recommendations for future research on gloss translation.

\begin{table}
    \centering
    \small
    \begin{tabular}{p{0.9\linewidth}}
      \toprule
      \textbf{Glosses (DSGS)} \\
      \texttt{KINDER FREUEN WARUM FERIEN NÄHER-KOMMEN} \\
      \midrule
      \textbf{Translation (DE)} \\
      \texttt{Die Kinder freuen sich, weil die Ferien näher rücken.} \\
      \midrule
      \textbf{Glosses (EN)} \\
      \texttt{(`CHILDREN REJOICE WHY HOLIDAYS APPROACHING')} \\
      \midrule
      \textbf{Translation (EN)} \\
      \texttt{(`The children are happy because the holidays are approaching.')} \\
      \bottomrule
    \end{tabular}
    \caption{Example for sign language glosses. DSGS=Swiss German Sign Language, DE=German, EN=English. English translations are provided for convenience. Example is adapted from a lexicon of the three sign languages of Switzerland, where a sign language video of this sentence is available (\url{https://signsuisse.sgb-fss.ch/de/lexikon/g/ferien/}).}
    \label{tbl:gloss_example}
\end{table}



Our paper makes the following contributions:

\begin{itemize}
    \item We provide a review of recent works on gloss translation (\S\ref{sec:related_work}).
    \item We outline recommendations for future work which promote awareness of the inherent limitations of gloss-based approaches, realistic datasets, stronger baselines and convincing evaluation (\S\ref{sec:best_practices_for_gloss_translation}).
\end{itemize}

\section{Related work}
\label{sec:related_work}


\begin{table*}
    \centering
    \tiny
    \begin{tabular}{l c cc ccc c cccc l}
      \toprule
      \multicolumn{1}{c}{} & 
      \multicolumn{1}{c}{\textbf{L}} & 
      \multicolumn{2}{c}{\textbf{datasets}} & \multicolumn{3}{c}{\textbf{translation directions}} & \multicolumn{1}{c}{\textbf{code}} & \multicolumn{4}{c}{\textbf{evaluation metrics}} & 
      \multicolumn{1}{c}{\textbf{BLEU tool}} \\
      \cmidrule(lr){2-2}
      \cmidrule(lr){3-4}
      \cmidrule(lr){5-7}
      \cmidrule(lr){8-8}
      \cmidrule(lr){9-12}
      \cmidrule(lr){13-13}
      & & \textbf{P} & \textbf{O} & \textbf{DGS$\rightarrow$DE} & \textbf{DE$\rightarrow$DGS} & \textbf{O} & & \textbf{B 1-3} & \textbf{B-4} & \textbf{R} & \textbf{O} \\
      
      \midrule
      
      
      \citet{8578910} & - & \ding{52} & - & \ding{52} & - & - & \href{https://github.com/neccam/nslt}{\ding{52}} & \ding{52} & \ding{52} & \ding{52} & - & \href{https://github.com/tensorflow/nmt/blob/master/nmt/scripts/bleu.py}{Tensorflow}
       \\
       
       \midrule
      
      
      \citet{stoll2018sign} & - & \ding{52} & - & - & \ding{52} & - & - & \ding{52} & \ding{52} & \ding{52} & - & (unclear) \\
      
      \midrule
      
      \citet{camgoz2020sign} & \ding{52} & \ding{52} & - & \ding{52} & - & - & \href{https://github.com/neccam/slt}{\ding{52}} & \ding{52} & \ding{52} & - & WER & (unclear)
       \\
      
      \midrule
      
      \citet{camgoz2020b} & \ding{52} & \ding{52} & - & \ding{52} & - & - & - & - & \ding{52} & \ding{52} & - & (unclear)
       \\
      
      \midrule
      
      \citet{yin-read-2020-better} & \ding{52} & \ding{52} & ASLG-PC12 & \ding{52} & - & ASL$\rightarrow$EN & \href{https://github.com/kayoyin/transformer-slt}{\ding{52}} & \ding{52} & \ding{52} & \ding{52} & METEOR & \href{https://github.com/kayoyin/transformer-slt/blob/master/tools/bleu.py}{NLTK} \\
      
      \midrule
      
      \citet{saunders2020progressive} & \ding{52} & \ding{52} & - & - & \ding{52} & - & \href{https://github.com/BenSaunders27/ProgressiveTransformersSLP}{\ding{52}} & \ding{52} & \ding{52} & \ding{52} & - & (unclear)
       \\
      
      \midrule
      
      \citet{stoll2020text2sign} & \ding{52} & \ding{52} & - & - & \ding{52} & - & - & \ding{52} & \ding{52} & \ding{52} & WER & (unclear)  \\
      
      \midrule
      
      \citet{orba-akarun} & - & \ding{52} & - & \ding{52} & - & - & - & \ding{52} & \ding{52} & \ding{52} & - & (unclear) \\
      
      \midrule
      
      \citet{moryossef-etal-2021-data} & - & \ding{52} & NCSLGR & \ding{52} & - & ASL$\rightarrow$EN & (\ding{52}) & - & \ding{52} & - & COMET & \href{https://github.com/mjpost/sacrebleu}{SacreBLEU} \\
      
      \midrule
      
      \citet{zhang-duh-2021-approaching} & - & \ding{52} & - & \ding{52} & \ding{52} & - & - & - & \ding{52} & - & - & (unclear) \\
      
      \midrule
      
      \citet{egea-gomez-etal-2021-syntax} & - & \ding{52} & - & - & \ding{52} & - & \href{https://github.com/LaSTUS-TALN-UPF/Syntax-Aware-Transformer-Text2Gloss}{\ding{52}} & - & \ding{52} & \ding{52} & METEOR, TER & \href{https://github.com/LaSTUS-TALN-UPF/Syntax-Aware-Transformer-Text2Gloss/blob/main/utils/sacreBLEU_script.py}{SacreBLEU}  \\
      
      \midrule
      
      \citet{saunders2021signing} & - & \ding{52} & DGS Corpus & - & \ding{52} & - & - & - & \ding{52} & \ding{52} & - & (unclear) \\
      
      \midrule
      
      \citet{angelova-etal-2022-using} & \ding{52} & \ding{52} & DGS Corpus & \ding{52} & - & - & \href{https://github.com/DFKI-SignLanguage/gloss-to-text-sign-language-translation}{\ding{52}} & - & \ding{52} & - & - & SacreBLEU \\
      
      \midrule
      
      \citet{walsh-saunders-bowden:2022:SLTAT} & - & \ding{52} & DGS Corpus & - & \ding{52} & - & - & \ding{52} & \ding{52} & \ding{52} & - &  (unclear) \\
      
      \bottomrule
    \end{tabular}
    \caption{Review of recent works on gloss translation. L=whether a paper discusses limitations of gloss approaches, P=\texttt{RWTH-PHOENIX-Weather 2014T} introduced by \citet{8578910}, O=other, (\ding{52})=exact shell commands are listed in the appendix of the paper, B=BLEU, R=ROUGE, B1-4=four variants of BLEU, varying the maximum ngram order from 1 to 4, SacreBLEU=version 1.4.14, COMET=\href{https://unbabel.github.io/COMET/html/models.html}{wmt-large-da-estimator-1719}, DGS=German Sign Language, DE=German, ASL=American Sign Language, EN=English. In the code column, checkmark symbols (\ding{52}) are clickable links.}
    \label{tbl:literature_review}
\end{table*}




For a general overview of sign language processing in the context of NLP see \citet{yin-etal-2021-including,moryossef2021slp} and \citet{de2022machine} for a comprehensive survey of sign language machine translation (including, but not limited to, gloss-based approaches).

We conduct a more narrow literature review of 14 recent publications on gloss translation. We report characteristics such as the datasets used, translation directions, and evaluation details (Table \ref{tbl:literature_review}).


\subsection{Awareness of limitations of gloss approach}
\label{subsec:awareness_of_limitations_of_gloss_approach}


We find that 8 out of 14 reviewed works do not include an adequate discussion of the limitations of gloss approaches, inadvertently overstating the potential usefulness of their experiments.

In the context of sign languages,
glosses are unique identifiers for individual signs. However, a linear sequence of glosses is not an adequate representation of a signed utterance, where different channels (manual and non-manual) are engaged simultaneously. Linguistically relevant cues such as non-manual movement or use of three-dimensional space may be missing \citep{yin-etal-2021-including}.

The gloss transcription conventions of different corpora vary greatly, as does the level of detail (see \citet{kopfmariaAnnotations2022} for an overview of differences and commonalities between corpora).
Therefore, glosses in different corpora or across languages are not comparable. Gloss transcription is an enormously laborious process done by expert linguists.

Besides, glosses are a linguistic tool, not a writing system established in Deaf
\footnote{It is a widely recognized convention to use the uppercased word \emph{Deaf} for describing members of a sign language community and, in contrast, to use the lowercased word \emph{deaf} when describing the audiological state of a hearing loss \citep{morgan-woll-2002}. More recent works \citep{napier-leeson-2016,kusters-et-al-2017} propose to use \emph{deaf} in both situations.}
communities. Sign language users generally do not read or write glosses in their everyday lives.

Taken together, this means that gloss translation suffers from an inherent and irrecoverable information loss,
that creating an abundance of translations transcribed as glosses is unrealistic, and that gloss translation systems are not immediately useful to end users.

\subsection{Choice of dataset}
\label{subsec:choice_of_dataset}

All reviewed works use the \texttt{RWTH-PHOENIX Weather 2014T} (hereafter abbreviated as \textit{PHOENIX}) dataset \citep{forster-etal-2014-extensions,8578910} while other datasets are used far less frequently. Besides, we note a distinct paucity of languages and translation directions: 12 out of 14 works are concerned only with translation between German Sign Language (DGS) and German (DE), the language pair of the PHOENIX dataset.

While PHOENIX was a breakthrough when it was published, it is of limited use for current research. 
The dataset is small (8k sentence pairs) and contains only weather reports, covering a very narrow linguistic domain. It is important to discuss the exact nature of glosses, how the corpus was created and how it is distributed.

\paragraph{Glossing}
PHOENIX is based on German weather reports interpreted into DGS and broadcast on the TV station Phoenix.
The broadcast videos served as input for the DGS side of the parallel corpus. Compared to the glossing conventions of other well-known corpora, PHOENIX glosses are simplistic and capture mostly manual features (with mouthings as the only non-manual activity covered), which is not sufficient to represent meaning (\S \ref{subsec:awareness_of_limitations_of_gloss_approach}).

\paragraph{Live interpretation and translationese effects} The fact that PHOENIX data comes from interpretation in a live setting has two implications: Firstly, since information was conveyed at high speed, the sign language interpreters omitted pieces of information from time to time. This leads to an information mismatch between some German sentences and their DGS counterparts. Secondly, due to the high speed of transmission, the (hearing) interpreters sometimes followed the grammar of German more closely than that of DGS. As a consequence, some translations are more similar to Signed German (\textit{Lautsprachbegleitendes Gebärden}) than DGS. 

\paragraph{Preprocessing of spoken language} The German side of the PHOENIX corpus is available only already tokenized, lowercased and with punctuation symbols removed. From an MT perspective this is unexpected since corpora are usually distributed without such preprocessing.

PHOENIX is popular because it is freely available and is a benchmark with clearly defined data splits introduced by \citet{8578910}. Sign language machine translation as a field is experiencing a shortage of free and open datasets and, with the exception of PHOENIX, there are no agreed-upon data splits.


Essentially, from a scientific point of view achieving higher gloss translation quality on the PHOENIX dataset is near meaningless. The apparent overuse of PHOENIX is reminiscent of the overuse of MNIST \citep{lecun2010mnist} in machine learning, or the overuse of the WMT 14 English-German benchmark in the MT community, popularized by \citet{DBLP:journals/corr/VaswaniSPUJGKP17}.

\subsection{Evaluation}
\label{subsec:evaluation}

As evaluation metrics, all works use some variant of BLEU \citep{papineni-etal-2002-bleu}, and ten out of 14 use some variant of ROUGE \citep{lin-2004-rouge}. All but four papers do not contain enough information about how exactly BLEU was computed. Different BLEU implementations, settings (e.g. ngram orders, tokenization schemes) and versions are used.


\begin{table*}
    \centering
    \small
    \begin{tabular}{lp{10cm}}
      \toprule
      \textbf{Reference} & \texttt{VIEL1A FAMILIE1* JUNG1 FAMILIE1 GERN1* IN1* HAMBURG1* STADT2* WOHNUNG2B* FAMILIE1} \\
      \textbf{Hypothesis} & \texttt{VIEL1B JUNG1 LEBEN1 GERN1* HAMBURG1* STADT2* \$INDEX1} \\
      \midrule
      \textbf{BLEU with tokenization} & 25.61 \\
      \textbf{BLEU without tokenization} & 10.18 \\
      \bottomrule
    \end{tabular}
    \caption{Impact of applying or disabling internal tokenization (mtv13a) when computing BLEU on gloss outputs. Example taken from the Public DGS Corpus \citep{hanke:20016:sign-lang:lrec}.}
    \label{tbl:bleu_tokenization_difference}
\end{table*}

\paragraph{Non-standard metrics} ROUGE is a metric common in automatic summarization but not in MT, and was never correlated with human judgement in a large study.
In eight out of 14 papers, BLEU is used with a non-standard maximum ngram order, producing variants such as BLEU-1, BLEU-2, etc. Similar to ROUGE, these variants of BLEU have never been validated as metrics of translation quality, and their use is scientifically unmotivated.


\paragraph{Tokenization}

BLEU requires tokenized machine translations and references. Modern tools therefore apply a tokenization procedure internally and implicitly (independently of the MT system's preprocessing).
Computing BLEU with tokenization on glosses leads to seemingly better scores but is misleading since tokenization creates many trivial matches.
For instance, in corpora that make use of the character \texttt{\$} in glosses (e.g. the DGS Corpus \citep{konrad_reiner_2022_10251}), \texttt{\$} is split off as a single character, inflating the ngram sub-scores.
For an illustration see Table \ref{tbl:bleu_tokenization_difference} (and Appendix \ref{appendix:impact_of_internal_tokenization} for a complete code listing) where we demonstrate that using or omitting tokenization leads to a difference of 15 BLEU.

\paragraph{Spurious gains}

Different implementations of BLEU or different tokenizations lead to differences in BLEU bigger than what many papers describe as an ``improvement'' over previous work \citep{post-2018-call}. Incorrectly attributing such improvements to, for instance, changes to the model architecture amounts to a ``failure to identify the sources of empirical gains'' \citep{lipton-steinhardt}.
In a similar vein, we observe that papers on gloss translation tend to copy scores from previous papers without knowing whether the evaluation procedures are in fact the same. This constitutes a general trend in recent MT literature \citep{marie-etal-2021-scientific}.

In summary, some previous works on gloss translation have used 1) automatic metrics that are not suitable for MT or 2)
well-established MT metrics in ways that are not recommended.
BLEU with standard settings and tools is inappropriate for gloss outputs.

The recommended way to compute BLEU on gloss output is to use the tool SacreBLEU \citep{post-2018-call} and to disable internal tokenization.
Nevertheless, even with these precautions, it is important to note that BLEU was never validated empirically as an evaluation metric for gloss output. Some aspects of BLEU may not be adequate for a sequence of glosses, such as its emphasis on whitespaces to mark the boundaries of meaningful units that are the basis of the final score.

Other string-based metrics such as CHRF \citep{popovic-2016-chrf} may be viable alternatives for gloss evaluation. CHRF is a character-based metric and its correlation with human judgement is at least as good as BLEU's \citep{kocmi-etal-2021-ship}.

\subsection{Further observations}
\label{subsec:further_observations}

More informally (beyond what we show in Table \ref{tbl:literature_review}), we observe that most papers do not process glosses in any corpus-specific way, that particular modeling and training decisions may not be ideal for low-resource gloss translation and that document-level systems may be crucial.


\paragraph{Preprocessing glosses}

Glosses are created for linguistic purposes (\S \ref{subsec:awareness_of_limitations_of_gloss_approach}), not necessarily with machine translation in mind. Particular gloss parts are not relevant for translation and, if kept, make the problem harder unnecessarily.
For instance, a corpus transcription and annotation scheme might prescribe that meaning-equivalent, minor form  variants of signs\footnote{In spoken language linguistics, these occurrences are called \emph{allophonic variants}. We refrain from using the term here, as sign languages do not consist of sounds (\emph{phones}).} are transcribed as different glosses.
An example of meaning-equivalent form variants in some sign languages is a flat handshape with the thumb adducted (aligned with the remaining fingers) or spread.
%
Reading or producing such meaning-equivalent variants may not be relevant for an MT system and makes the learning problem harder.

Since the particular nature of glosses is specific to every corpus, it is necessary to preprocess glosses in a corpus-specific way. We illustrate corpus-specific gloss processing in Appendix \ref{appendix:example_for_corpus_specific_gloss_preprocessing}, using the Public DGS Corpus \cite{hanke:20016:sign-lang:lrec} as an example.
 

\paragraph{Modeling and training decisions}


Gloss translation experiments are certainly low-resource scenarios and therefore, best practices for optimizing MT systems on low-resource datasets apply \citep{sennrich-zhang-2019-revisiting}. For example, dropout rates or label smoothing should be set accordingly, and the vocabulary of subwords of a subword model should be generally small \citep{ding-etal-2019-call}.


Gloss translation models are often compared to other approaches as baselines, it is therefore problematic if those gloss baselines are weak and unoptimized \citep{denkowski-neubig-2017-stronger}.

\paragraph{Limitations of sentence-level systems}

All surveyed works propose sentence-level systems, as opposed to larger-context models.
For sign language MT specifically, sentence-level approaches have inherent limitations.

Signed languages exhibit frequent referencing (or ``indexing'') behaviour similar to anaphora in spoken languages. But while in spoken languages ambiguities arising from anaphora can often be resolved with sentence-level context, that is not the case for signed languages. After a subject was introduced with a sign once it is often unacceptable to use the sign again in a later utterance of the discourse, instead the subject is referred to with an index (see e.g. \citealp{engberg-pedersen_1993}). Besides, sign languages are generally not gendered (which for spoken languages helps disambiguate by narrowing down
possible antecedents).


Document-level benchmarks (e.g. \citealp{bawden-etal-2018-evaluating,muller-etal-2018-large}) show that for spoken languages, document-level context is not crucial since sentence-level baseline systems have relatively high performance. For signed languages, discourse context may be indispensable.

We are not aware of any existing discourse-level translation system for sign languages but we believe they could be investigated in the future.

\section{Recommendations for gloss translation}
\label{sec:best_practices_for_gloss_translation}

Based on our review of recent works on gloss translation, we make the following recommendations for future research:

\begin{itemize}
    \item Demonstrate awareness of limitations of gloss approaches (\S\ref{subsec:awareness_of_limitations_of_gloss_approach}), and explicitly discuss them in the paper.
    \item Focus on datasets beyond PHOENIX. Openly discuss the limited size and linguistic domain of PHOENIX (\S\ref{subsec:choice_of_dataset}).
    \item Use metrics that are well-established in MT. Compute BLEU with SacreBLEU, report metric signatures and disable internal tokenization for gloss outputs. Do not compare to scores produced with a different or unknown evaluation procedure (\S\ref{subsec:evaluation}).
    \item Given that glossing is corpus-specific (\S\ref{subsec:awareness_of_limitations_of_gloss_approach}), process glosses in a corpus-specific way, informed by the respective transcription conventions (\S\ref{subsec:further_observations}).
    \item Optimize gloss translation baselines with methods shown to be effective for low-resource MT (\S\ref{subsec:further_observations}).
\end{itemize}

We also believe publishing reproducible code makes works on gloss translation more valuable. 

\section{Conclusion}

In this paper we have shown that some recent works on gloss translation lack awareness of the inherent limitations of glosses and common datasets, as well as a standardized evaluation method (\S\ref{sec:related_work}). In order to make future research on gloss translation more meaningful, we make practical recommendations (\S\ref{sec:best_practices_for_gloss_translation}).

We urge researchers to spell out limitations of gloss translation approaches, e.g. in the now mandatory limitation sections of *ACL papers, and to strengthen their findings by implementing existing best practices in MT.

Finally, we also caution that researchers should consider whether gloss translation is worthwhile, and if time and effort would be better spent on basic linguistic tools (such as segmentation, alignment or coreference resolution), creating training corpora or translation methods that do not rely on glosses.











\section*{Data licensing}

The license of the Public DGS Corpus\footnote{\url{https://www.sign-lang.uni-hamburg.de/meinedgs/ling/license_en.html}} (which we use only as examples in Table \ref{tbl:bleu_tokenization_difference} and Appendix \ref{appendix:example_for_corpus_specific_gloss_preprocessing}) does not allow any computational research except if express permission is given by the University of Hamburg.

\section*{Acknowledgements}

This work was funded by the EU Horizon 2020 project EASIER (grant agreement no. 101016982) and the Swiss Innovation Agency  (Innosuisse) flagship IICT (PFFS-21-47).

We thank the DGS Corpus team at the University of Hamburg for helpful discussions on gloss preprocessing.

\bibliography{custom}
\bibliographystyle{acl_natbib}

\clearpage
\onecolumn

\appendix

\section{Impact of internal tokenization when computing BLEU on gloss sequences}
\label{appendix:impact_of_internal_tokenization}

\begin{listing}[!ht]
\begin{minted}[linenos, frame=lines, xleftmargin=20pt, fontsize=\small]{python}

# ! pip install sacrebleu==2.2.0

>>> from sacrebleu.metrics import BLEU

# English translation: Many young families like living in the city of Hamburg.
# German translation: Viele junge Familien leben gerne in Hamburg in der Stadt.

>>> ref = "VIEL1A FAMILIE1* JUNG1 FAMILIE1 GERN1* IN1* HAMBURG1* STADT2* WOHNUNG2B* FAMILIE1"

>>> hyp = "VIEL1B JUNG1 LEBEN1 GERN1* HAMBURG1* STADT2* $INDEX1"

# computing BLEU on gloss output with tokenization (not recommended):

>>> bleu = BLEU() # default: BLEU(tokenize="13a")
>>> bleu.corpus_score([hyp], [[ref]])
BLEU = 25.61 63.6/50.0/33.3/25.0 (BP = 0.635 ratio = 0.688 hyp_len = 11 ref_len = 16)

# computing BLEU on gloss output without tokenization (recommended):

>>> bleu = BLEU(tokenize="none")
>>> bleu.corpus_score([hyp], [[ref]])
BLEU = 10.18 57.1/16.7/10.0/6.2 (BP = 0.651 ratio = 0.700 hyp_len = 7 ref_len = 10)

\end{minted}
\caption{Impact of enabling or disabling internal tokenization (mtv13a) when computing BLEU on gloss outputs.}
\label{listing:impact_of_incorrectly_applying_internal_tokenization}
\end{listing}

\section{Example for corpus-specific gloss preprocessing}
\label{appendix:example_for_corpus_specific_gloss_preprocessing}

For this example we recommend to download and process release 3.0 of the corpus.
To DGS glosses we suggest to apply the following modifications derived from the DGS Corpus transcription conventions \citep{konrad_reiner_2022_10251}:

\begin{itemize}
    \item Removing entirely two specific gloss types that cannot possibly help the translation: \texttt{\$GEST-OFF} and \texttt{\$\$EXTRA-LING-MAN}.
    \item Removing \textit{ad-hoc} deviations from citation forms, marked by \texttt{*}. Example: \texttt{ANDERS1*} $\rightarrow$ \texttt{ANDERS1}.
    \item Removing the distinction between type glosses and subtype glosses, marked by \texttt{\^{}}. Example: \texttt{WISSEN2B\^{}} $\rightarrow$ \texttt{WISSEN2B}. 
    \item Collapsing phonological variations of the same type that are meaning-equivalent. Such variants are marked with uppercase letter suffixes. Example: \texttt{WISSEN2B} $\rightarrow$ \texttt{WISSEN2}.
    \item Deliberately keep numerals (\texttt{\$NUM}), list glosses (\texttt{\$LIST}) and finger alphabet (\texttt{\$ALPHA}) intact, except for removing handshape variants.
\end{itemize}

See Table \ref{tbl:gloss_preprocessing_examples} for examples for this preprocessing step. Overall these simplifications should reduce the number of observed forms while not affecting the machine translation task. For other purposes such as linguistic analysis our preprocessing would of course be detrimental.

\newpage

\begin{table*}
    \centering
    \small
    \begin{tabular}{lp{13cm}}
      \toprule
      \textbf{before} & \texttt{\$INDEX1 ENDE1\^{} ANDERS1* SEHEN1 MÜNCHEN1B* BEREICH1A*} \\
      \textbf{after} & \texttt{\$INDEX1 ENDE1 ANDERS1 SEHEN1 MÜNCHEN1 BEREICH1} \\
      \midrule
      \textbf{before} & \texttt{ICH1 ETWAS-PLANEN-UND-UMSETZEN1 SELBST1A* KLAPPT1* \$GEST-OFF\^{} BIS-JETZT1 GEWOHNHEIT1* \$GEST-OFF\^{}*} \\
      \textbf{after} & \texttt{ICH1 ETWAS-PLANEN-UND-UMSETZEN1 SELBST1 KLAPPT1 BIS-JETZT1 GEWOHNHEIT1} \\
      \bottomrule
    \end{tabular}
    \caption{Examples for preprocessing of DGS glosses.}
    \label{tbl:gloss_preprocessing_examples}
\end{table*}

While this preprocessing method provides a good baseline, it can certainly be refined further. For instance, the treatment of two-handed signs could be improved. If a gloss occurs simultaneously on both hands, we either keep both glosses or remove one occurrence. In both cases, information about the simultaneity of signs is lost during preprocessing and preserving it could potentially improve translation.

\end{document}